\documentclass[conference]{IEEEtran}
\IEEEoverridecommandlockouts
\usepackage{cite}
\usepackage{layout}
\usepackage{array}
\usepackage{threeparttable}
\usepackage{booktabs}
\usepackage{multirow}
\usepackage{multicol}
\usepackage{amsmath}
\usepackage{url}
\usepackage{hyperref}
\usepackage{amsmath,amssymb,amsfonts}
\usepackage{algorithmic}
\usepackage{graphicx}
\usepackage{textcomp}
\usepackage{xcolor}
\def\BibTeX{{\rm B\kern-.05em{\sc i\kern-.025em b}\kern-.08em
    T\kern-.1667em\lower.7ex\hbox{E}\kern-.125emX}}
\begin{document}

\title{HMPDM: A Diffusion Model for Driving Video Prediction with Historical Motion Priors\\

}

\author{Ke Li, ~\IEEEmembership{Graduate Student Member,~IEEE},
Tianjia Yang, 
Kaidi Liang,
Xianbiao Hu, 
Ruwen Qin\IEEEauthorrefmark{1},~\IEEEmembership{Member,~IEEE}

\thanks{Ke Li, Kaidi Liang, and Ruwen Qin are with the Department of Civil Engineering, Stony Brook University, Stony Brook, NY 11794, USA.}
\thanks{Tianjia Yang and Xianbiao Hu are with the Department of Civil and Environmental Engineering, Pennylvania State University, University Park, PA 16802, USA.}
\thanks{\IEEEauthorrefmark{1} Corresponding author: Ruwen Qin, email: ruwen.qin@stonybrook.edu}
}
\maketitle

\begin{abstract}
Video prediction is a useful function for autonomous driving, enabling intelligent vehicles to reliably anticipate how driving scenes will evolve and thereby supporting reasoning and safer planning. However, existing models are constrained by multi-stage training pipelines and remain insufficient in modeling the diverse motion patterns in real driving scenes, leading to degraded temporal consistency and visual quality. To address these challenges, this paper introduces the historical motion priors-informed diffusion model (HMPDM), a video prediction model that leverages historical motion priors to enhance  motion understanding and temporal coherence. The proposed deep learning system introduces three key designs: \textit{(i)} a Temporal-aware Latent Conditioning (TaLC) module for implicit historical motion injection; \textit{(ii)} a Motion-aware Pyramid Encoder (MaPE) for multi-scale motion representation; \textit{(iii)} a Self-Conditioning (SC) strategy for stable iterative denoising. Extensive experiments on the Cityscapes and KITTI benchmarks demonstrate that HMPDM outperforms state-of-the-art video prediction methods with efficiency, achieving a 28.2\% improvement in FVD on Cityscapes under the same monocular RGB input configuration setting. The implementation codes are publicly available at \url{https://github.com/KELISBU/HMPDM}.
\end{abstract}


\section{Introduction}
In the context of intelligent vehicles, modern autonomous driving systems need to not only perceive the present scenarios \cite{percieve} but also  anticipate their evolution over time, which is crucial for path planning, especially in safety-critical scenarios \cite{planning}. However, the complex dynamics and frequent occlusions in real-world driving scenes cause traditional object-centric predictors, which are based on low-dimensional states (e.g., trajectories), to discard essential appearance, structural, and contextual information. Emerging diffusion-based video prediction methods aim to forecast entire future scenes conditioned on past or present visual context. Crucially, scene-level video prediction offers a comprehensive and holistic understanding of the driving environment, capturing both the static background and the motion of dynamic traffic agents. Furthermore, by leveraging historical motion patterns, they naturally handle occlusions and can better represent the dynamic evolution of driving scene into the future. 

Despite notable advances in diffusion-based video prediction, several key challenges remain. Many models struggle with temporal consistency, since the visual quality degrades and objects appear distorted within the long-horizon prediction. Another limitation lies in the lack of robust historical motion modeling. Without properly leveraging historical motion priors, predicted agents often exhibit unrealistic trajectories or blurry motion patterns in driving scenarios. Accompanied with it is the concern of computational efficiency and adaptability. Despite promising performance, recent models often require substantial computation, complex training pipelines, and extra multimodal inputs, revealing the need for simpler yet effective solutions for driving video generation.

To bridge the gaps, the paper proposes a diffusion-based driving video prediction framework that is aware of historical motion, named \textbf{H}istorical \textbf{M}otion \textbf{P}riors-informed \textbf{D}iffusion \textbf{M}odel (\textbf{HMPDM}), making the following contributions: 
\begin{itemize}
    \item {To implicitly inject historical motion context, \textbf{T}emporal-\textbf{a}ware \textbf{L}atent \textbf{C}onditioning (\textbf{TaLC}) is introduced to feed the model with latent representations of past frames as a learnable prior.}
    \item {Complementing this, a \textbf{M}otion-\textbf{a}ware \textbf{P}yramid \textbf{E}ncoder (\textbf{MaPE}) hierarchically encodes multi-scale motion features from the historical dynamic.}
    \item {To mitigate error accumulation in long-term generation, we implement \textbf{S}elf-\textbf{C}onditioning (\textbf{SC}), a strategy where the model conditions on its own intermediate predictions.}
\end{itemize}

The remainder of this paper is organized as follows. Sec.~\ref{sec:related work} reviews the relevant literature. Sec.~\ref{sec:methodology} presents the proposed HMPDM framework in detail. The experimental setting and results are reported in Sec.~\ref{sec:exp and results}. Finally, Sec.~\ref{sec:conclusion} concludes the study and discusses future directions to pursue.

\section{Related Work}
\label{sec:related work}


This study is built upon the literature on diffusion-based video prediction, historical motion injection, and motion-enhanced video prediction.

\begin{figure*}[t]
\vspace{2mm}
\centering
\includegraphics[width=0.9\linewidth]{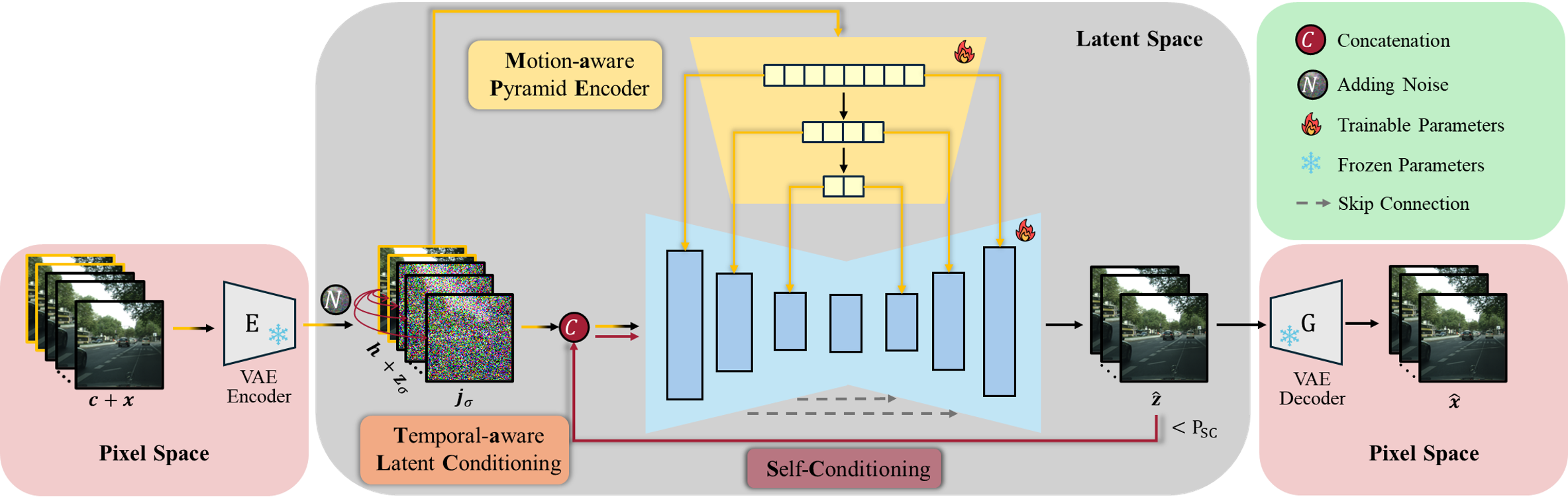}
\caption{Overview of the proposed HMPDM framework} \label{fig:methodology}
\end{figure*}

\subsection{Diffusion-based Video Prediction}
A stream of studies primarily relies on LSTM, VAEs, and VRNNs \cite{LSTM1,LSTM2,VRNN,GHVAEs,SVG} to learn a probabilistic latent space, model temporal dynamics, and capture spatial coherence. However, they meet the limitation on long-term video prediction with temporal consistency. Recently, diffusion-based paradigms have demonstrated remarkable performance in both video generation and prediction tasks. A diffusion model for video prediction is a generative framework that progressively transforms random Gaussian noise into coherent videos through an iterative denoising process \cite{DDPM,DDIM,EDM}. Many studies build upon the U-Net denoising architecture, incorporating various conditioning strategies and spatio-temporal consistency techniques \cite{VDM,RVD,RaMViD,MCVD,STDIFF,vdim,Syncvp}.

\subsection{Historical Motion Injection}
Historical frames naturally provide motion priors as the conditional context for future video prediction. Common ways for incorporating historical frames either concatenate them on the frame or channel dimension, or first encode them and subsequently inject the encoded representations into the cross-attention layers.
NPVP \cite{ye2023unified} used a CNN autoencoder to extract appearance features from past frames, which are used as keys and values in the cross-attention layers to predict future frames. Similarly, STDiff \cite{STDIFF} leveraged difference images from past frames as input to a specialized motion encoder, which disentangles motion and content features. While, VDT \cite{vdt} compared three different historical motion injection schemes and illustrated the effectiveness of direct token concatenation. Integrating both injection methods, LGC-VD \cite{LGC-VD} introduced a two-stage training design, where recent local frames are concatenated on the channel dimension, while longer-range global history is encoded and injected via cross-attention mechanism. However, these methods struggle to effectively unify local and global motion priors while maintaining alignment across scales.

\subsection{Motion-enhanced Multimodal Video Prediction}
To enhance time coherence and motion consistency of video prediction, recent approaches have incorporated multimodal data, such as depth, optical flow, and contour as complementary information. ExtDM \cite{ExtDM} and LFDM \cite{LFDM} designed motion autoencoders to extract flow-based motion information to guide the video diffusion process. Additionally, Syncvp \cite{Syncvp} incorporated conditional depth information with RGB video within a synchronous denoising framework. Whereas, most of them rely on a two-stage training pipeline to handle multimodal inputs, thereby increasing computational cost and complicating cross-modal alignment.

\begin{figure*}[t]
\centering
\vspace{1mm}
\includegraphics[width=0.7\linewidth]{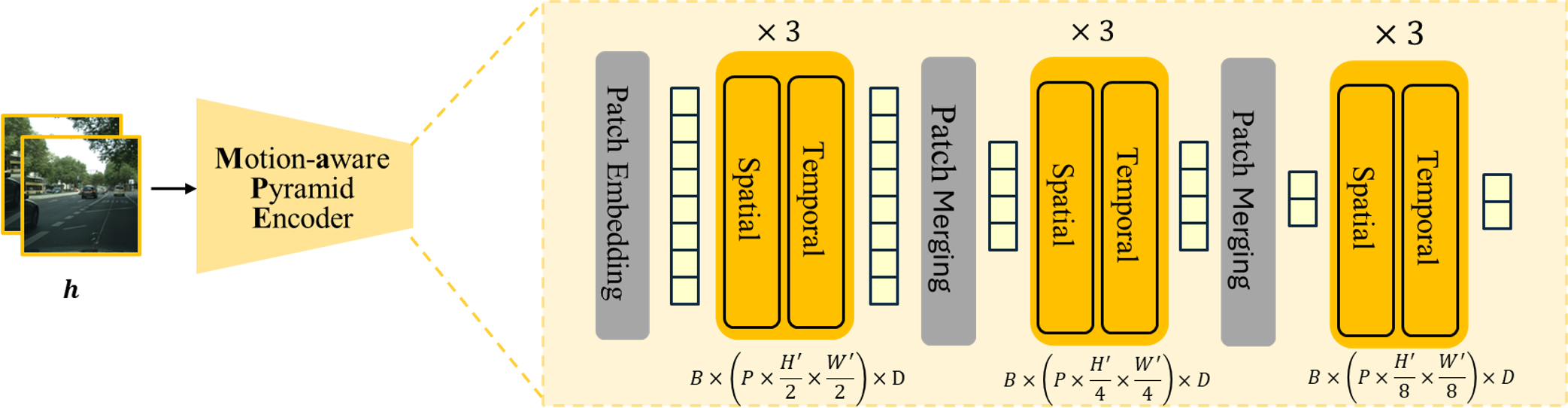}
\caption{MaPE architecture} \label{fig:MaPE}
\end{figure*}

\section{Methodology}
\label{sec:methodology}
We propose a simple, yet effective, diffusion model for driving video prediction that uses RGB cameras and exploits historical motion priors to enhance the quality of generated data. The overall framework of the proposed method is illustrated in Fig. \ref{fig:methodology}. HMPDM aims to generate and predict $F$ future video frames, $ \pmb{x}=\{\pmb{x}_{t}\}_{t=1}^{F}$, given a set of $P$ past frames, $\pmb{c}= \{\pmb{c}_{t}\}_{t=1}^{P}$, where $\pmb{x}_{t}$ and $\pmb{c}_{t}$ $\in \mathbb{R}^{C \times H \times W}$ are RGB images. Therefore, the objective is to learn the conditional distribution of future frames $p(\pmb{x}|\pmb{c})$. The proposed framework comprises three main components: temporal-aware latent conditioning,  a motion-aware pyramid encoder, and a self-conditioning strategy.

\subsection{Temporal-aware Latent Conditioning}
\label{subsec:Temporal-aware Latent Conditioning}
HMPDM builds upon the latent diffusion modeling paradigm, specifically following the Elucidated Diffusion Model (EDM) framework originally proposed by Karras et al.\cite{EDM} and widely adopted in Stable Video Diffusion (SVD) \cite{svd}. The latent autoencoder $E$ encodes a sequence of
past frames $\pmb{c}\in \mathbb{R}^{B \times P\times C\times H \times W}$ and future frames $\pmb{x}\in \mathbb{R}^{B \times F\times C\times H \times W}$ to a low-dimensional representations, $\pmb{h}$ and $\pmb{z}$, denoted as:
\begin{equation}
    \begin{aligned}
        \pmb{h}=E(\pmb{c}), \quad \pmb{z}=E(\pmb{x}),
    \end{aligned}
\end{equation}
where $\pmb{h} \in \mathbb{R}^{B \times P\times C' \times H' \times W'}$, $\pmb{z}\in \mathbb{R}^{B \times F\times C' \times H' \times W'}$, and $B$ denotes the batch size. As demonstrated in Vista \cite{Vista}, integrating the clean latent representations of past frames $\pmb{h}$ with noisy latent representations of future frames  $\pmb{z}_\sigma$ does not degrade generation quality. The forward diffusion process of HMPDM progressively corrupts the clean latent $\pmb{z}$ by adding noise with scale $\sigma \in [\sigma_{\min}, \sigma_{\max}]$:
\begin{equation}
\pmb{z}_\sigma = \pmb{z} + \sigma \pmb{n}, \quad \pmb{n} \sim \mathcal{N}(0, I),
\end{equation}
where $\sigma$ controls the corruption level. The resulting noisy latent is conditionally Gaussian: $
\pmb{z}_\sigma \mid \pmb{z} \sim \mathcal{N}(\pmb{z},\, \sigma^2 I)
$. A joint input, $\pmb{j}_\sigma$, is defined as:
\begin{equation}
    \pmb{j}_\sigma=\operatorname{Concat(\pmb{h},\pmb{z}_\sigma)},
\end{equation} 
which implicitly injects temporal context into the U-Net $\mathcal{E}$, allowing its internal spatio-temporal attention layers to jointly model and attend to the historical conditioning information. 

To differentiate the deterministic nature of the observed domain from the stochasticity of the noisy domain, the time embedding for clean past frames and noisy future latent are defined as:
\begin{equation}
\label{eq:embed}
    \begin{aligned}
        &e_{\text{clean}}(\sigma) = \operatorname{Embed_{clean}}(0.25\log\sigma_\mathrm{min}),\\
        &e_{\text{noise}}(\sigma) = \operatorname{Embed_{noise}}(0.25\log \sigma),
    \end{aligned}
\end{equation}
where $\operatorname{{Embed}_{clean}}(\cdot)$ and $\operatorname{Embed_{noise}}(\cdot)$ share the same architecture, a sinusoidal time embedding function, and are both initialized with the same pretrained weights. Then, we construct a binary mask $\pmb{m}\in \{0,1\}^{(P+F)}$, where the first $P$ positions (corresponding to historical frames) are set to 1, and the remaining $F$ positions (future frames) are set to 0. This allows for defining the frame-wise time embedding as:
\begin{equation}
\label{eq:joint Embed}
    e(\sigma)=\pmb{m}\cdot e_\text{clean}(\sigma) + (1-\pmb{m})e_\text{noise}(\sigma).
\end{equation}
Eqs. (\ref{eq:embed}) and (\ref{eq:joint Embed}) means that $\pmb{e}(\sigma)$ enforces a stationary, minimal noise level for historical frames, effectively distinguishing the clean and noisy domain to enhance the diffusion process's awareness of motion observed from past frames.

To predict the clean signal $\hat{\pmb{x}}$ from the joint input $\pmb{j}_\sigma$ conditioned on $\pmb{h}$, the denoiser $\mathcal{D}$ is utilized:
\begin{equation}
\label{eq:denoiser}
\begin{aligned}
    \mathcal{D}_(\pmb{j}_\sigma; \sigma,\pmb{h}) =& c_{\text{skip}}(\sigma)\pmb{j}_\sigma \\
    &+ c_{\text{out}}(\sigma)\mathcal{E}(c_{\text{in}}(\sigma)\pmb{j}_\sigma; e(\sigma), \pmb{h}),
\end{aligned}
\end{equation}
where $e(\cdot)$ is the time embedding defined in \eqref{eq:joint Embed}, and $c_{\text{skip}}$, $c_{\text{in}}$, and $c_{\text{out}}$ are scale-dependent coefficients for normalization and conditioning:
\begin{equation}
\label{eqa:embed}
\begin{aligned}
&c_{\text{skip}}(\sigma)   =1/(\sigma^2 + 1), \\
&c_{\text{in}}(\sigma)   = \sqrt{c_{\text{skip}}(\sigma)^2},\; c_{\text{out}}(\sigma)    =\sqrt{1-c_{\text{in}}(\sigma)^2 }.
\end{aligned}
\end{equation}
$\mathcal{E}$ in Eq. (\ref{eq:denoiser})  is the diffusion model's encoder to be introduced in Sec. \ref{subsec: Motion-aware Pyramid Encoder}, where the mechanism of explicitly using historical latent frames $\pmb{h}$ to enhance video generation is introduced.

\subsection{Motion-aware Pyramid Encoder}
\label{subsec: Motion-aware Pyramid Encoder}

The Motion-aware Pyramid Encoder (MaPE), shown in Fig. \ref{fig:MaPE}, is a hierarchical spatio-temporal transformer encoder that convert the latent representation of historical video frames $\pmb{h}$ into multi-scale token sequences. Following the design of SVD \cite{svd}, $\mathcal{E}$ adopts a 3D U-Net architecture composed of down, mid, and up blocks of spatial and temporal layers:
\begin{equation}
\mathcal{E}=\operatorname{Up}(\operatorname{Mid}(\operatorname{Down}(\pmb{j}_\sigma;\sigma,\pmb{h}))).
\end{equation}
$\pmb{M}_s$, for $s=1,2,3$, are token sequences produced by MaPE. Those tokens capture both local and global historical motion priors injected into $\mathcal{E}$ via cross-attention layers. 
$\pmb{M}_s$ ($\in \mathbb{R}^{B\times N_s \times D}$) consists of $N_s$ tokens, whose spatial token grids align with the corresponding grids used in the intermediate outputs of $\mathcal{E}$. As illustrated in Fig. \ref{fig:MaPE}, the framework of MaPE is composed of patch embedding module for tokenization, patch merging modules for pyramid downsampling, and three sequential blocks, each consisting of multiple spatio-temporal attention layers for capturing motion dynamics. 

Concretely, patch embedding module divides the latent representation of historical frames, $\pmb{h}$, into non-overlapping $2\times2$ patches and linearly projects them into a $D$-dimensional  embedding space through a learnable patch embedding layer, where $D$ is consistent with the hidden dimension of cross-attention layers in the U-Net. To better extract inner context and motion dynamics of historical condition cues, a stack of alternating spatio-temporal transformer blocks \cite{latte} is utilized. Driven by them, fine-grained local tokens $\pmb{M}_1$ with short-term motion cues are denoted as:
\begin{equation}
    \pmb{M}_1=(\operatorname{Spatial}\circ\operatorname{Temporal})^3(\operatorname{PatchEmbed}(\pmb{h})),
\end{equation}
where $\pmb{M}_1 \in \mathbb{R}^{B \times N_1\times D}$ is produced by 3 alternating spatio-temporal transformer blocks, and $N_1=P\times\frac{H'}{4} \times \frac{W'}{4}$ . Constructing a hierarchical feature pyramid by performing progressive $2 \times 2$ patch merging across stages facilitates the learning of structural and object-centric motion representation in $\pmb{M}_2$, 
\begin{equation}
    \pmb{M}_2=(\operatorname{Spatial}\circ\operatorname{Temporal})^3(\operatorname{PatchMerg}(\pmb{M}_1)),
\end{equation}
where  $\pmb{M}_2 \in \mathbb{R}^{B \times N_2\times D}$ and $N_2=P\cdot \frac{H'}{4} \cdot \frac{W'}{4}$ tokens.
Global semantic with long-term temporal dependencies are in $\pmb{M}_3$,
\begin{equation}
    \pmb{M}_3=(\operatorname{Spatial}\circ\operatorname{Temporal})^3(\operatorname{PatchMerg}(\pmb{M}_2)),
\end{equation}
where $\pmb{M}_3 \in \mathbb{R}^{B \times N_3\times D}$ and $N_3=P\cdot \frac{H'}{8} \cdot \frac{W'}{8}$. 

These motion-aware tokens $[\pmb{M}_1,\pmb{M}_2,\pmb{M}_3]$ are employed as conditional memory in U-Net through cross-attention layers at different depth. At each scale $s\in\{1,2,3\}$, the hidden state $\pmb{Z}_s$ from each U-Net block serves as a sequence of query tokens, while the token sequences $\pmb{M}_s$ act as the key-value memory. The attention operation at scale $s$ is formulated as:
\begin{equation}
\begin{gathered}
Q_s = \pmb{W}_Q \pmb{Z}_s,\quad
K_s = \pmb{W}_K \pmb{M}_s,\quad
V_s = \pmb{W}_V \pmb{M}_s, \\
\operatorname{Attn}_s =
\operatorname{Softmax}(
    Q_s K_s^{\top}/\sqrt{d}
),
\end{gathered}
\end{equation}
where $\pmb{W}_Q$, $\pmb{W}_K$, and $\pmb{W}_V\in \mathbb{R}^{D\times d}$ are learnable matrices that project the query, key, and value token sequences into the attention subspace of dimension $d$. 

The diffusion model then learns to reverse this process by training with the denoising score matching objective:
\begin{equation}
\mathcal{L}_{\text{diff}}(\theta) =
\mathbb{E}_{\substack{
\pmb{z}, \pmb{h}\sim p_\mathrm{data}, \sigma \sim p(\sigma),\\
\pmb{z}_\sigma \mid \pmb{z} \sim \mathcal{N}(\pmb{z},\, \sigma^2 I)
}}
\left[ \lambda_{\sigma}\, \| \mathcal{D}_\theta(\pmb{j}_\sigma;\sigma,\pmb{h}) - \pmb{z} \|_2^2 \right],
\end{equation}
where $\lambda_{\sigma}$ balances noise levels and encourages prediction robustness across different $\sigma$ values, \(p_\mathrm{data}\) denotes the data distribution of clean video frames in latent space, and $p(\sigma)$ is defined as a discrete uniform distribution followed by EDM.

\begin{table*}[t]
\centering
\begin{threeparttable}
\caption{Quantitative comparison of video prediction models on Cityscapes and KITTI datasets. $\downarrow$ means lower is better, $\uparrow$ means higher is better}
\begin{tabular}{>{\raggedright\arraybackslash}p{0.9in}rlc|rccc|ccc}
\toprule
\multirow{2}{*}{\textbf{Methods}} & \multirow{2}{*}{\textbf{Year}}& \multirow{2}{*}{\textbf{Input}} &\multirow{2}{*}{\textbf{Pipeline}}  &
\multicolumn{4}{c|}{\textbf{Cityscapes}(128$\times$128) 2 $\rightarrow$ 28} & \multicolumn{3}{c}{\textbf{KITTI}(128$\times$128) 4 $\rightarrow$ 5} \\
\cmidrule(lr){5-8} \cmidrule(lr){9-11}
 & & & &\textbf{FVD} $\downarrow$ & \textbf{SSIM} $\uparrow$ & \textbf{PSNR} $\uparrow$ & \textbf{LPIPS} $\downarrow$
  & \textbf{SSIM} $\uparrow$ & \textbf{PSNR} $\uparrow$ & \textbf{LPIPS} $\downarrow$ \\
\midrule
\multicolumn{4}{l|}{\textbf{256 Random Samples}}& &  & &  &  &  &    \\

U-ViT \cite{U-VIT}       & CVPR23 &R &1  &1045.3 & 0.362 & 10.84 & 0.431  & -- & -- & -- \\
RaMViD \cite{RaMViD}       & TMLR24 &R &1  &812.6  & 0.454 & 13.14 & 0.395  & -- & -- & -- \\
VDIM \cite{vdim}       & AAAI23 &R &2  &724.7  & 0.539 & 18.49 & 0.252  & -- & -- & -- \\
RVD \cite{RVD}         & ArXiv22 &R &1  &465.0  & 0.489 & 17.21 & 0.226  & -- & -- & -- \\
MCVD-s$^{\ast}$ \cite{MCVD}      & NeurIPS22 &R &1   &184.8 & \underline{0.720} & \underline{22.50} &\underline{0.121}  & -- & -- & -- \\
LFDM \cite{LFDM}       & CVPR23 &R+F  &2  &194.9  & 0.601 & 20.32 & 0.157  & -- & -- & -- \\
ExtDM-K4 \cite{ExtDM} & CVPR24 &R+F &2  & \textbf{121.3} & \textbf{0.745} & \textbf{22.84} &  \textbf{0.108}  & -- & -- & -- \\
\textbf{HMPDM (Ours)$^{\ast}$}        & IV26 &R & 1&   \underline{151.2}  &0.633 & 21.42 &0.142  & -- & -- & -- \\
\midrule
\multicolumn{4}{l|}{\textbf{Full Test Set}}& &  & &  &  &  &    \\
NPVP \cite{ye2023unified}       & CVPR23  &R & 2  &768.0  & \textbf{0.744} &  --   & 0.183  &\textbf{0.66} & -- & 0.279 \\
LGC-VD \cite{LGC-VD}     &IJCAI23 & R & 2& 124.6 & \underline{0.732}  & -- & \textbf{0.069} & -- & --& --  \\
STDiff$^{\ast}$ \cite{STDIFF}      &AAAI24 & R &1 & 107.3 & 0.658  & -- & \underline{0.136}  & 0.54 & --& \textbf{0.115}  \\
SyncVP$^{\ast}$ \cite{Syncvp}      &CVPR25 & R+D & 2& \underline{84.0} & 0.649  & -- & 0.160 & -- & --& --  \\
\textbf{HMPDM (Ours)$^{\ast}$}         &IV26  & R&1 & \textbf{77.0}  & 0.626 & \textbf{21.37} & 0.145 & \underline{0.54} & \textbf{18.62} & \underline{0.149} \\
\bottomrule

\end{tabular}

\begin{tablenotes}\footnotesize
\item[$\ast$] Results computed with \#T=10; unmarked entries use \#T=100 or the original paper's default is unspecified.
\item[] R = RGB; F = optical flow; D = depth.
\end{tablenotes}
\end{threeparttable}
\label{tab:quantitative_comparison}
\end{table*}

\subsection{Self-conditioning}
\label{subsec: Self Conditioning}
During inference, conditioning each denoising step on the model’s previous generation, named self-conditioning, allows the model to review its historical trajectories and thus enhance motion-aware temporal consistency \cite{selfconditin1,selfcondition2}. Therefore, in the training stage, with probability $p_\mathrm{sc}$, a forward pass is first performed without gradient updates. Then, the detached prediction is concatenated to the input along the channel dimension for a second forward pass on which gradients are computed. With probability $1-p_{sc}$, we follow the default of SVD, the most recent historical frame is replicated across the temporal sequence and concatenated along the channel dimension. The empirical findings reported in W.A.L.T. \cite{selfcondition2} suggest setting $p_\mathrm{sc}$ as 0.9. 

Specifically, following an $n$-step discrete noise schedule, the noise level $\sigma_i$ decreases gradually,  for $i=n,\dots,1$. At the beginning of the sampling process ($i=n$), a sequence of latent variables is initiated from a Gaussian distribution with variance $\sigma_{\max}^2$. Given conditioning frames $\pmb{h}$, the denoiser then produces an estimate of the clean latent at each step. For the remaining steps ($i=n-1,\dots, 1$), the latent variables are iteratively refined using the discrete EDM update, and the previous estimated variables are concatenated along channel dimension as the next input. After $n$ denoising steps, the final clean future latent $\hat{\pmb{z}}$ is decoded into RGB frames $\hat{\pmb{x}}$ by a VAE decoder $G$, as follows:
\begin{equation}    \hat{\pmb{x}}=G(\hat{\pmb{z}}).
\end{equation}

\section{Experiments and Results}
\label{sec:exp and results}

\subsection{Implementation Details}
\subsubsection{Setting}

The proposed HMPDM was implemented using PyTorch 1.10.0 on a server equipped with an Nvidia L40S featuring 48 GB of memory. The diffusion model is trained for 10$^5$ steps using AdamW optimizer, with a batch size ($B$) of 4 and a learning rate $2\times10^{-5}$. The frozen VAE and the trainable U-Net are initialized with the pretrained weight from SVD \cite{svd}.

\subsubsection{Datasets}
To evaluate the effectiveness of the proposed framework for driving video prediction, experiments are conducted on Cityscapes \cite{cityscapes} and KITTI \cite{kitti} datasets. Following the standard video prediction protocol, both datasets are resized to $128 \times 128$. Specifically, the Cityscapes dataset is partitioned into 2,975 training clips, 500 validation clips, and 1,525 test clips, each containing 30 consecutive frames. The KITTI dataset is divided into 759 training clips and 150 test clips, with each clip containing 9 frames.

\subsubsection{Evaluation Metrics}
Following the protocols established in prior work~\cite{MCVD,Syncvp,ExtDM}, four commonly used metrics are adopted to evaluate the performance of video prediction models: Structural Similarity Index Measure (SSIM) \cite{ssim}, Peak Signal-to-Noise Ratio (PSNR), Learned Perceptual Image Patch Similarity (LPIPS) \cite{lpips}, and Fréchet Video Distance (FVD) \cite{FVD}. SSIM and PSNR evaluate pixel-wise reconstruction fidelity, while LPIPS measures perceptual similarity in deep feature space. FVD captures the spatio-temporal consistency by quantifying both motion dynamics and temporal coherence. For fair comparison, all metrics are reported on both a randomly sampled subset of 256 clips and the full test set across 10 random future denoising trajectories (\#T), from which the best performing result is selected. To compare computational complexity, we additionally report the input modality, and the number of stages in the training pipeline.

\subsection{Comparison with State-of-the-art Models}

To evaluate the effectiveness of the proposed HMPDM on driving video prediction, we benchmark it against SOTA models on Cityscapes and KITTI datasets following the two standard protocols, with results summarized in Table \ref{tab:quantitative_comparison}. Under a strictly RGB-only input setting, HMPDM demonstrates superior performance on Cityscapes. Specifically, it achieves an FVD of 151.2 on the 256-sample test protocol, representing a 17.8\% relative reduction compared to MCVD \cite{MCVD}, and 77.0 on the full-test protocol, corresponding to a 28.2\% relative reduction to STDiff \cite{STDIFF}. Although HMPDM is limited to a single RGB modality, it attains competitive results even against multimodal models (e.g., R+F, R+D), ranking the second on the 256-sample test set while achieving the best FVD on the full test set. Furthermore, HMPDM retains a one-stage training pipeline and requires fewer optimization steps, emphasizing its efficiency and compact architecture.

In contrast to FVD, HMPDM's performances on SSIM, PSNR, and LPIPS illustrate comparable yet slightly inferior results to the SOTA models. This disparity primarily stems from the inference mechanism. These evaluation metrics are calculated per frame, whereas HMPDM performs a one-time prediction of the entire future segment. Conversely, SOTA methods \cite{ExtDM,MCVD,LGC-VD,Syncvp} adopt autoregressive rollouts with shorter prediction horizons, which attenuate the drift, even in cases where the long-term motion consistency metric FVD is not superior.

To further assess the generalization of HMPDM in autonomous driving scenarios, we evaluate on KITTI with a short prediction horizon (4$\rightarrow$5). Given the limited number of predicted frames, FVD is omitted for this dataset and we only report SSIM, PSNR, and LPIPS. The performance of high PSNR and competitive SSIM indicate strong pixel-level alignment and high structural similarity, while the LPIPS gap relative to the best method suggests remaining room in texture reconstruction. Notably, since HMPDM learns and samples in the VAE latent space at a low dimensional space (128 $\times$ 128), high-frequency details may be smoothed, contributing to its modest deficit on LPIPS.

\begin{figure*}[!t]
\vspace{1mm}
\centering
\includegraphics[width=1.05\columnwidth]{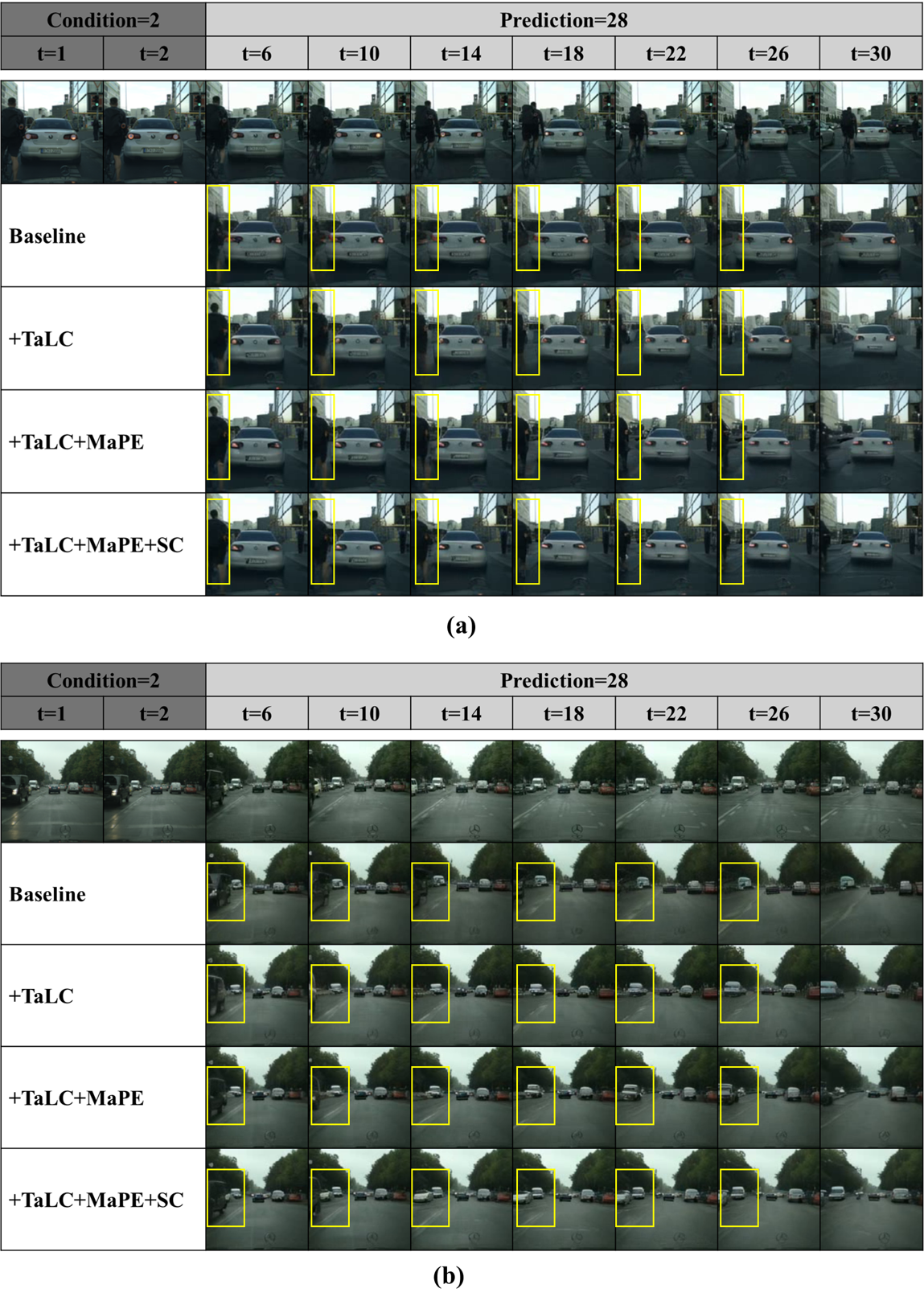}
\caption{Qualitative Ablation Results on Cityscapes (128$\times$ 128)}
\label{fig:ablation1}
\end{figure*}

\subsection{Ablation Study}
To assess the ability of the HMPDM framework in modeling historical motion priors and video prediction, experiments are conducted on the Cityscapes dataset. Sec. \ref{subsubsec:Effectiveness of Components} quantifies contribution of each component within HMPDM, while Sec. \ref{subsubsec:Effectiveness of Conditioning Horizon} examines how the conditioning horizon influences prediction quality.

\subsubsection{Effectiveness of Components}
\label{subsubsec:Effectiveness of Components}

\begin{table}[t]
\centering
\caption{Quantitative Ablation Results on Cityscapes. $\downarrow$ means lower is better, $\uparrow$ means higher is better}
\begin{tabular}{l|cccc}
\toprule
\multirow{2}{*}{\textbf{Models}} &
\multicolumn{4}{c}{\textbf{Cityscapes}(128$\times$128) 2 $\rightarrow$ 28} \\
\cmidrule(lr){2-5} 
 &  \textbf{FVD} $\downarrow$ & \textbf{SSIM} $\uparrow$ & \textbf{PSNR} $\uparrow$ & \textbf{LPIPS} $\downarrow$
  \\
\midrule
Baseline        & 236.2 & 0.574 & 19.94 & 0.193 \\
$+$ TaLC       & 197.5  & 0.627 & 21.42 & 0.153 \\
$+$ TaLC $+$ MP         & 188.7  & 0.632 &  \textbf{21.52}   & 0.154 \\
$+$ TaLC $+$ MaPE  & \underline{155.7}  &\textbf{0.634} &21.44  & \underline{0.146}  \\
$+$ TaLC $+$ MaPE $+$ SC & \textbf{151.2}  &\underline{0.633} & 21.42 & \textbf{0.142}  \\
\bottomrule
\end{tabular}
\label{tab:ablation1}
\end{table}

Table \ref{tab:ablation1} summarizes the contribution of each component for predicting 28 future frames conditioned on 2 past frames. The architecture and pretrained weights of baseline model are identical to those of SVD \cite{svd} and then fine-tuned on the Cityscapes dataset. Compared to this baseline model, adding the TaLC module yields 16.4\% lower FVD, 9.2\% higher SSIM, 7.4\% higher PSNR, and 21\% lower LPIPS. These improvements indicate that the TaLC effectively leverages implicit historical context, leading to improved motion dynamics and stronger structural fidelity.

We further evaluate the MaPE module with respect to its pyramid design and patch merging mechanism. First, the variant (+TaLC+MP) improves over the variant (+TaLC) by 4.5\% in FVD, highlighting the importance of multi-scale conditioning injected through cross-attention layers for capturing the semantically relevant features. +TaLC+MaPE, which integrates past frames within the spatio-temporal domain through stacked transformers, achieves a 17.5\% reduction in FVD compared with the variant (+TaLC+MP), illustrating the effectiveness of the designed transformer blocks.  

Finally, adding SC upon the variant (+TaLC+MaPE) constitutes the complete HMPDM framework, which delivers the best FVD and LPIPS and maintains competitive SSIM and PSNR. This strategy mitigates error accumulation and enhances temporal consistency. 

The contribution of each designed component is further visualized in Fig. \ref{fig:ablation1}. Qualitative comparisons demonstrate that the TaLC enhances the spatio-temporal consistency, as evidenced by improved spatial structure. Specifically, in both samples, the front vehicle occupies lager spatial area in the frames generated by the baseline model than in the corresponding ground truth frames. Furthermore, the addition of MaPE allows HMPDM to better capture the historical motion priors. Compared with predicted frames generated by (+TaLC) in sample (a), the (+TaLC+MaPE) variant produces clearer motion patterns for the cyclist, as evidenced by more visually coherent leg movements. Integrating the variant (+TaLC+MaPE) with SC, further enables HMPDM to handle occlusions and preserve the fidelity of traffic agents. As shown in sample (b), HMPDM successfully generates the stationary white sedan even though it never appears in the past frames. These improvements enhance the quality of generated traffic videos, thus providing more realistic and reliable information to support vehicles in predicting and reasoning about future scenes.

\subsubsection{Effectiveness of Conditioning Horizon}
\label{subsubsec:Effectiveness of Conditioning Horizon}

Intuitively, providing sufficient historical motion information enhances the dynamic priors required for future frame generation. To investigate the impact of varying conditioning horizons on future driving video prediction, quantitative results on the Cityscapes dataset are reported in Table \ref{tab:ablation2}. When increasing the number of historical frames from 2 to 6 while keeping the same prediction length, the FVD decreases from 117.4 to 104.9 ($\approx$10.7\%$\downarrow$).  Simultaneously, SSIM improves from 0.679 to 0.694, PSNR increases from 22.92 to 23.37 dB, and LPIPS concurrently decreases from 0.117 to 0.110. These improvements indicate that supplying richer motion context advances both temporal coherence and frame-wise fidelity, with diminishing but still positive returns as the conditioning length grows.
\begin{table}[t]
\centering
\caption{Effectiveness of Conditioning Horizon on Video Prediction Performance. $\downarrow$ means lower is better, $\uparrow$ means higher is better}
\begin{tabular}{l|cccc}
\toprule
\multirow{2}{*}{\textbf{Conditioning Horizon}} &
\multicolumn{4}{c}{\textbf{Cityscapes}(128$\times$128)} \\
\cmidrule(lr){2-5} 
 &  \textbf{FVD} $\downarrow$ & \textbf{SSIM} $\uparrow$ & \textbf{PSNR} $\uparrow$ & \textbf{LPIPS} $\downarrow$
  \\
\midrule
2 $\rightarrow$ 14        & 117.4 & 0.679 & 22.92 & 0.117 \\
4 $\rightarrow$ 14       & 108.9  & 0.691 & 23.24 & 0.112 \\
6 $\rightarrow$ 14         & 104.9  & 0.694 &  23.37   & 0.110 \\
\bottomrule
\end{tabular}
\label{tab:ablation2}
\end{table}
\section{Conclusion}
\label{sec:conclusion}

This paper proposes HMPDM, a diffusion-based video prediction model enhanced by mono-modal historical motion priors and specifically tailored for driving scenarios. TaLC and MaPE modules effectively capture and utilize the historical motion priors from past driving patterns. In addition, SC improves the fidelity of traffic agents and long-term temporal consistency. The proposed HMPDM framework, owing to its efficiency and well-designed historical motion priors, achieves superior performance on the Cityscapes benchmark, outperforming existing methods by 17.8\% and 28.2\% in FVD under two standard evaluation protocols, respectively.

The HMPDM lays a foundation for the perception and prediction in emerging driving world models. Future work will explore efficient multimodal extensions and conduct deployment-oriented evaluations to improve robustness for real-world utilization.
\section*{Acknowledgment}
Qin receives funding (69A3552348321) from the Rural Safe Efficient Advanced Transportation Center (R-SEAT), funded by the US Department of Transportation (USDOT). The contents of this paper reflect the views of the authors. USDOT assumes no liability for the contents or use thereof.

\bibliographystyle{IEEEtran} 
\bibliography{IEEEexample}

\end{document}